\pdfoutput=1

\documentclass[11pt]{article}

\usepackage[]{EMNLP2024}

\usepackage{times}
\usepackage{latexsym}

\usepackage[T1]{fontenc}

\usepackage[utf8]{inputenc}

\usepackage{microtype}

\usepackage{inconsolata}

\usepackage{hyperref}
\usepackage{url}
\usepackage{multirow}
\usepackage{bbm}
\usepackage{graphicx}
\usepackage{amssymb}
\usepackage{amsmath} 
\usepackage{booktabs}
\usepackage{lscape}
\usepackage{pifont}
\usepackage{algorithm}
\usepackage{graphicx}
\usepackage{caption}
\usepackage{subcaption}
\usepackage{CJKutf8}
\usepackage{algorithm}
\usepackage{algpseudocode}

\newcommand\benchmarkname{\textcolor{black}{\textsc{MetaMetrics-MT}}}

\title{$\benchmarkname$: Tuning Meta-Metrics for Machine Translation via Human Preference Calibration}

\author{David Anugraha$^\dagger$$^1$, Garry Kuwanto$^\dagger$$^2$, Lucky Susanto$^3$,\\
\textbf{Derry Tanti Wijaya$^{\ddagger, 2,3}$, Genta Indra Winata}$^\ddagger$\thanks{$\text{ }$ The work was conducted outside Capital One. $^\dagger$These authors contributed equally. $^\ddagger$Senior authors.}\hspace{0.17cm}$^4$ \\
  $^1$University of Toronto $\quad$ $^2$Boston University\\
  $^3$Monash Indonesia $\quad$ $^4$Capital One \\
  \texttt{david.anugraha@cs.toronto.edu, \{gkuwanto,wijaya\}@bu.edu,}\\
  \texttt{lucky.susanto@monash.edu, genta.winata@capitalone.com}}

\begin{document}
\maketitle
\begin{abstract}
We present $\benchmarkname$, an innovative metric designed to evaluate machine translation (MT) tasks by aligning closely with human preferences through Bayesian optimization with Gaussian Processes. $\benchmarkname$ enhances existing MT metrics by optimizing their correlation with human judgments. Our experiments on the WMT24 metric shared task dataset demonstrate that $\benchmarkname$ outperforms all existing baselines, setting a new benchmark for state-of-the-art performance in the reference-based setting. Furthermore, it achieves comparable results to leading metrics in the reference-free setting, offering greater efficiency.
\end{abstract}

\section{Introduction}

Evaluating machine translation (MT) tasks is inherently complex, as no single metric can universally apply to all scenarios. A metric that performs well for one task may not be suitable for another, and its effectiveness can vary significantly depending on the specific language pairs involved. Therefore, relying solely on a single metric is often inadequate. To ensure the usefulness of automatic metrics, it is crucial to align them with human annotations~\cite{winata2024preference}. To achieve a more comprehensive evaluation, benchmarks typically incorporate multiple metrics, such as lexical-based and semantic-based metrics. However, the correlation between these metrics can be skewed due to variations in the models used and the training data employed for evaluation. For instance, BERTScore~\cite{zhang2019bertscore} uses contextual embeddings from pre-trained transformers to assess performance, with different models excelling in specific language pairs. In contrast, neural-based metrics like BLEURT \cite{sellam2020bleurt}, COMET \cite{rei2020comet}, and CometKiwi \cite{rei2022comet} employ distinct methodologies and training datasets. 
These differences can affect each metric's alignment with human judgments and their reliability across language pairs. Some metrics, like XCOMET-Ensemble \cite{guerreiro2023xcomet}, demand high memory (at least 80GB), prompting efforts to predict LLM performance using smaller models \cite{anugraha2024proxylm}.

In this paper, we propose $\benchmarkname$, a MT metric inspired by $\textsc{MetaMetrics}$~\cite{winata2024metametrics}. This meta-metric is crafted to align more closely with human preferences through the use of Bayesian optimization with Gaussian Processes (GP). By systematically integrating multiple existing metrics, $\benchmarkname$ achieves state-of-the-art performance for reference-based metrics and shows a strong correlation with human scores for reference-free metrics in the WMT24 metric shared task~\cite{metrics2024}. Through the strategic combination of metrics with assigned weights, $\benchmarkname$ aims to be as competitive as, if not superior to, any individual metric. Our contributions include the following:
\begin{itemize}
    \item We present $\benchmarkname$ in reference-based and reference-free settings, offering flexibility for various MT scenarios. Our reference-based model sets the state-of-the-art for the WMT24 task. We publicly release the code for easy usability.\footnote{The code is available at \url{https://github.com/meta-metrics/metametrics}.}
    \item We demonstrate that the $\benchmarkname$ metric is easily adjustable to meet the human preference. 
    \item We show that $\benchmarkname$ is compact and efficient, capable of running on a commercial GPU with 40GB of memory, whereas a comparable metric like XCOMET-Ensemble requires significantly higher memory with at least 80GB.
\end{itemize}

\section{Methodology}

\subsection {\benchmarkname}
$\benchmarkname$ is designed to leverage multiple metrics for assessing MT tasks, with each metric being adjusted by specific weights to optimize performance. The idea of utilizing multiple metrics is to combine scores from multiple metrics regardless of the metric types. Formally, let $\theta_{1}, \theta_{2}, \ldots, \theta_{N}$ represent $N$ distinct metric functions with $\hat{y}_{1}, \ldots, \hat{y}_{N}$ as their respective performance on a translation task. We define $\Phi$ to compute a scalar meta-metric score of $\hat{y}_\textsc{MM}$ using $\hat{y}_{1}, \ldots, \hat{y}_{N}$. Overall, we define $\theta_{\textsc{MM}}$ as a meta-metric function where $\hat{y}_\textsc{MM}$ is computed as follows:
\begin{align}
    \hat{y}_i &= \theta_i(x), \\
    \quad \hat{y}_\textsc{MM} = \theta_{\textsc{MM}}(x) &= \Phi(\hat{y}_1, \cdots, \hat{y}_N).
\end{align}

Our objective is to calibrate a metric function, $\theta_\textsc{MM}$, to maximize the correlation $\rho(\hat{y}_\textsc{MM}, \gamma)$, where $\rho$ is a correlation measure and $\gamma$ represents human assessment scores, which include any scores provided by human evaluators. Each metric operates within a specific range, defined by minimum and maximum values. However, some metrics, particularly those based on neural networks, may fall outside this range. To ensure consistency, we normalize these metrics to a common scale from 0 to 1, where 0 signifies poor translation performance and 1 signifies perfect translation performance. In this process, given an original score $y_i$ for a given metric, $\tilde{y}_i$ represents the normalized score. For more details on pre-processing, please refer to Section~\ref{sec-preprocessing} of the Appendix.

In this case, we use GP to model the function $\Phi$ and it can be breakdown into a weighted sum as follows:
\begin{align}
y_\textsc{MM} = \alpha_{1} \tilde{y_{1}} + \alpha_{2} \tilde{y_{2}} + \ldots + \alpha_{N} \tilde{y_{N}},
\end{align}
where $\alpha_{1}, \alpha_{2}, \ldots, \alpha_{N}$ are the corresponding weights assigned to each metric, constrained to the interval $[0, 1]$. Our objective is to determine the best set of weights for $\alpha_{1}, \alpha_{2}, \ldots, \alpha_{N}$, which maximizes $\rho(y_\textsc{MM}, \gamma)$. Notice that $y_\textsc{MM}$ lies in the interval of $[0, N]$, so normalizing $y_\textsc{MM}$ back to $[0, 1]$ is unnecessary as linear scaling does not affect the correlation coefficient for correlation function $\rho$.

The advantage of $\benchmarkname$ is its flexibility and adaptability across tasks and domains. By integrating metrics that strongly correlate with human judgments for specific tasks, we can create a composite metric that improves overall alignment with human evaluations.
\begin{table}[!t]
\centering
\resizebox{0.49\textwidth}{!}{
    \begin{tabular}{lcccr}
    \toprule
    \textbf{Metric} & \textbf{clipping} & \textbf{normalization} & \textbf{inversion} & \textbf{weight} \\ \midrule
    \multicolumn{5}{l}{\textbf{Reference-based} ($\benchmarkname$)}\\ \midrule
    MetricX-23-XXL & [0,25] & $\checkmark$ & $\checkmark$ & 1.0000 \\
    COMET & [0,1] & $\checkmark$ & $\times$ & 0.2055 \\
    XCOMET-XL & [0,1] & $\checkmark$ & $\times$ & 0.2733 \\
    \midrule
    \multicolumn{5}{l}{\textbf{Reference-free} ($\benchmarkname$-QE)}\\ \midrule
    MetricX-23-XXL-QE & [0,25] & $\checkmark$ & $\checkmark$ & 0.9905 \\
    CometKiwi (QE) & [0,1] & $\checkmark$ & $\times$ & 0.1267\\
    CometKiwi-XL (QE) & [0,1] & $\checkmark$ & $\times$ & 0.0584 
    \\
    \bottomrule
    \end{tabular}
}
\caption{Metric configuration for $\benchmarkname$. Metrics not listed in the table have been assigned a weight of zero.}
\label{tab:best-configs}
\end{table}
\subsection{Bayesian Optimization}
We optimize the weights for each metric using Bayesian optimization with GP as the surrogate model. Bayesian optimization is particularly useful in this context because it efficiently explores and exploits the parameter space when the objective function is expensive to evaluate. By constructing a probabilistic model of the objective function, Bayesian optimization balances exploring new areas with exploiting known promising regions, making it effective even when evaluations are costly.

The GP constructs a joint probability distribution over the variables, assuming a multivariate Gaussian distribution. As the number of observations increases, the posterior distribution becomes more precise, enabling the algorithm to more effectively identify promising regions in the weight space. The Bayesian optimization process involves several iterations. First, the GP model is updated by fitting it to the observed data. Next, the algorithm selects the next set of weights by maximizing the acquisition function, which uses the posterior distribution to choose the next sample from the search space. Finally, the objective function is evaluated at these weights. This iterative process continues until a convergence criterion is met, ensuring that the optimization effectively identifies the optimal weights for the metrics.

\subsection{$\benchmarkname$ Settings}
\subsubsection{Hybrid Mode}
In the WMT24 shared task dataset, we observe that some samples lack references in the challenge sets, even for reference-based metrics. To address this issue, we implement a hybrid mode that switches from reference-based to reference-free metrics when reference data is unavailable.

\subsubsection{Same Language Optimization}
During the optimization process, we train a dedicated model for each known language pair in the training set to ensure optimal performance. If a language pair is not present in the training set, we use the entire dataset for tuning.

\section{Experimental Setup}

\subsection{Training Datasets and Hyper-parameters}

We introduce two versions of $\benchmarkname$ to accommodate both reference-based and reference-free evaluations: $\benchmarkname$, which employs reference-based metrics, and $\benchmarkname$-QE, which utilizes reference-free metrics. We train $\benchmarkname$ and $\benchmarkname$-QE using 3 years of MQM datasets from the WMT shared tasks spanning 2020 to 2022~\cite{mathur2020results,freitag2021results,freitag2022results}. The dataset used for tuning is at the segment level, with Kendall's $\tau$ correlation as the evaluation metric. For the Bayesian optimization, we run GP with a Matérn kernel \cite{williams2006gaussian}, a generalization of the RBF kernel, using $\nu = 2.5$. The optimization is performed over 100 steps, starting with 5 initialization points.

\subsection{Metrics for \benchmarkname}

We describe the reference-based metrics utilized for $\benchmarkname$. During the selection process, we included only metrics that can run on a commercial GPU with 40GB of memory. Consequently, XCOMET-XXL and CometKiwi-XXL were not considered. Additionally, we limited the use of the OpenAI API to \texttt{GPT4o-mini}, which is significantly more cost-effective than other GPT-4 model options. 

\subsubsection{Reference-based Metric}
We utilize nine different metrics in our optimization, including three variations of MetricX-23 and two different BERTScore metrics using precision and F1. The metrics under study are as follows:
\begin{table*}[!th]
\centering
\resizebox{\textwidth}{!}{
    \begin{tabular}{lcc|cccc|cccc|cccc}
    \toprule
    \textbf{Model} & \multicolumn{2}{c|}{\textbf{overall}} & \multicolumn{4}{|c|}{\textbf{en-de}} & \multicolumn{4}{|c|}{\textbf{en-es}}& \multicolumn{4}{|c}{\textbf{ja-zh}} \\ 
    & $r$ & sys/seg & $r$  &  sys & $r$ &  seg&  $r$  &sys     & $r$   & seg  & $r$   & sys & $r$ & seg \\
    & \multicolumn{2}{c|}{avg. corr} &  & SPA & & acc-t & & SPA & & acc-t & & SPA &  & acc-t \\ \midrule
    \textbf{Reference-based} \\ \midrule
    sentinel-ref-mqm & 10 & 0.513 & 7 & 0.405 & 18 & 0.429 &4& 0.581 & 8 & 0.680 &	8 & 0.545 & 17 & 0.435 \\
    BLEU & 9 & 0.589 & 4 & 0.736 & 16 & 0.431 & 6 & 0.512 & 	8 & 0.680 & 6 & 0.740 & 17 & 0.435 \\
    spBLEU & 9 & 0.593 & 4 & 0.741 & 17 & 0.431 & 6 & 0.523 & 7 & 0.680 & 6	& 0.744 & 16 & 0.436 \\
    chrfS & 8 & 0.606 &	4 & 0.742 & 14 & 0.434 & 6 & 0.549 & 6 & 0.682 & 4 & 0.788 & 14 & 0.444 \\
    chrF & 8 & 0.608 & 4 & 0.750 & 15 & 0.431 & 5 & 0.581 & 8 & 0.680 & 5 & 0.767 & 16 & 0.436 \\
    MEE4 & 7 & 0.609 & 5 & 0.731 & 13 & 0.437 & 7 & 0.504 & 	4 & 0.683 & 2	& 0.855 & 13 & 0.446 \\
    BERTScore & 7 & 0.617 &	4 & 0.749 &	14 & 0.435 & 4 & 0.587 & 6 & 0.682 & 4 & 0.799 & 12 & 0.451 \\
    YiSi-1 & 6 & 0.630 & 4 & 0.759 & 13 & 0.436 & 4 & 0.609 & 7 & 0.681 & 3 & 0.835 & 11 & 0.458 \\
    PrismRefSmall & 5 & 0.642 & 4 & 0.772 & 14 & 0.433 & 4 & 0.634 & 8 & 0.680 & 2 & 0.875 & 11 & 0.457 \\
    PrismRefMedium & 5 & 0.646 & 4 & 0.776 & 14 & 0.434 &	3 & 0.652 & 7 & 0.680 & 2 & 0.872 & 10 & 0.462 \\
    BLCOM\_1 & 4 & 0.664 & 3 & 0.840 & 10 & 0.455 & 3 & 0.680 & 6 & 0.681 & 3 & 0.843 & 7 & 0.488 \\
    BLEURT-20 & 3 & 0.686 & 2 & 0.881 & 7 & 0.486 & 3 & 0.695 & 6 & 0.681 & 1 & 0.887	& 8	& 0.484 \\
    COMET & 3 & 0.688 & 2 & 0.879 & 8	& 0.482 & 2	& 0.778 & 5 & 0.683 & 4 & 0.813 & 6 & 0.496 \\
    XCOMET & 2 & 0.719 & 1 & \textbf{0.906} & 3 & 0.530 & 2 & 0.788 & 1 & \textbf{0.688} & 2 & 0.890 & 7 & 0.510 \\
    MetricX-24 (Hybrid) & 1	& 0.721 & 2	& 0.874 & 2 & 0.532 & 2 & 0.799 & 3 & 0.685 & 1 & \underline{0.897} & 2 & 0.539 \\
    \midrule
    $\benchmarkname$ & 1	& \underline{0.724} & 2 & 0.882 & 1 & \textbf{0.542} & 2 & \textbf{0.804} & 2	& \underline{0.686} & 3 & 0.871 & 1 & \textbf{0.561} \\
    $\benchmarkname$ (Same Lang.) & 2 & 0.723 & 1 & \underline{0.883} & 1 & \textbf{0.542} & 2 & \underline{0.803} & 2 & \underline{0.686} & 3 & 0.874 & 2 & \underline{0.550} \\
    $\benchmarkname$ (Hybrid) & 1 & \textbf{0.725} & 2	& \underline{0.883} & 1	& \textbf{0.542}	& 1	& \textbf{0.804} & 2	& \underline{0.686}	& 2	& 0.873	& 1	& \textbf{0.561} \\
    \midrule
    \textbf{Reference-free}
    \\ \midrule
    CometKiwi & 5 & 0.640 & 5 & 0.732 & 9 & 0.467 & 3 & 0.693 & 4 & 0.684 & 5 & 0.776 & 7 & 0.490 \\
    sentinel-cand-mqm & 5 & 0.650 & 3 & 0.822 & 4 & 0.517 & 2 & 0.785 & 4 & 0.683 & 7 & 0.610 & 8 & 0.481 \\
    bright-qe & 4 & 0.681 & 3 & 0.816 & 6 & 0.500 & 2 & 0.792 & 1 & \textbf{0.689} & 4 & 0.805 & 8 & 0.484 \\
    XCOMET-\textsc{QE} & 3 & 0.695 & 1 & \textbf{0.889} & 4 & \underline{0.520} & 1 & 0.801 & 2 & \underline{0.687} & 4 & 0.808 & 10 & 0.463 \\
    CometKiwi-XXL & 3 & 0.703 & 3 & 0.839 & 9 & 0.481 & 1 & \textbf{0.843} & 8 & 0.680 & 2 & \underline{0.881} & 8 & 0.494 \\
    gemba\_esa & 2 & \underline{0.711} & 4 & 0.793 & 5 & 0.507 & 1 & \underline{0.838} & 5 & 0.683 & 1 & \textbf{0.908} & 2 & \textbf{0.539} \\
    MetricX-24-\textsc{QE} (Hybrid) & 2 & \textbf{0.714} & 2 & 0.878 & 3 & \textbf{0.526} & 2 & 0.789 & 4 & 0.685 & 2 & 0.875 & 3 & \underline{0.530} \\ \midrule
    $\benchmarkname$-\textsc{QE} & 3 & 0.684 & 2 & \underline{0.860} & 6 & 0.497 & 3 & 0.711 & 2 & 0.686 & 3 & 0.837 & 4 & 0.516 \\
    $\benchmarkname$-\textsc{QE} (Same Lang.) &  4 & 0.688 & 2 & \underline{0.860} & 7 & 0.497 & 4 & 0.709 & 2 & 0.686 & 4 & 0.853 & 5 & 0.524 \\
    \bottomrule
    \end{tabular}
}
\caption{WMT24 results (MQM). \textbf{Bold} and \underline{underline} values indicate the best and second best performance, respectively.}
\label{tab:qa-results}
\end{table*}

\begin{table}[ht]
    \centering
    \resizebox{0.49\textwidth}{!}{
    \begin{tabular}{lcccc}
        \toprule
        \textbf{Metric} & \textbf{all} & \textbf{en-de} & \textbf{en-es} & \textbf{ja-zh} \\
        \midrule
        \textbf{Reference-based} \\ \midrule
        sentinel-ref-mqm & 0.513 & 0.417 & 0.631 & 0.490 \\
        BLEU & 0.589 & 0.583 & 0.596 & 0.588 \\
        spBLEU & 0.593 & 0.586 & 0.602 & 0.590 \\
        chrF & 0.606 & 0.589 & 0.615 & 0.616 \\
        chrfS & 0.608 & 0.591 & 0.630 & 0.602 \\
        BERTScore & 0.610 & 0.584 & 0.594 & 0.651 \\
        MEE4 & 0.617 & 0.592 & 0.635 & 0.625 \\
        damonmonli & 0.640 & 0.599 & 0.688 & 0.633 \\
        YiSi-1 & 0.643 & 0.603 & 0.657 & 0.666 \\
        PrismRefSmall & 0.646 & 0.605 & 0.666 & 0.667 \\
        PrismRefMedium & 0.650 & 0.669 & 0.734 & 0.545 \\
        BLCOM\_1 & 0.684 & 0.679 & 0.698 & 0.676 \\
        BLEURT-20 & 0.686 & 0.683 & 0.688 & 0.685 \\
        COMET-22 & 0.695 & 0.705 & \underline{0.744} & 0.636 \\
        XCOMET & 0.719 & \textbf{0.717} & 0.740 & 0.700 \\ 
        MetricX-24 (Hybrid) & \underline{0.721} & 0.703 & 0.742 & \textbf{0.718} \\ \midrule
        $\benchmarkname$ (Hybrid) & \textbf{0.725} & \underline{0.713} & \textbf{0.745} & \underline{0.717} \\ \midrule
        \textbf{Reference-free} \\ \midrule
        sentinel-src-mqm & 0.513 & 0.418 & 0.630 & 0.491 \\
        XLsimMqm & 0.515 & 0.531 & 0.520 & 0.493 \\
        sentinel-cand-mqm & 0.630 & 0.597 & 0.645 & 0.647 \\
        CometKiwi & 0.635 & 0.569 & 0.644 & 0.691 \\
        bright-qe & 0.665 & 0.647 & 0.681 & 0.665 \\
        XCOMET-QE & 0.689 & \underline{0.680} & 0.730 & 0.655 \\
        MetricX-24-QE (Hybrid) & \textbf{0.714} & \textbf{0.702} & 0.737 & \underline{0.702} \\
        gemba\_esa & \underline{0.711} & 0.650 & \textbf{0.761} & \textbf{0.724} \\ \midrule
        $\benchmarkname$-\textsc{QE} & 0.681 & 0.658 & \underline{0.740} & 0.644 \\
        \bottomrule
    \end{tabular}
    }
\caption{Detailed WMT24 results per language category. \textbf{Bold} and \underline{underline} values indicate the best and second best performance, respectively.}
\label{tab:lang-cat-result}
\end{table}

\paragraph{BERTScore \cite{zhang2019bertscore}} The metric calculates cosine similarity scores for each token in the candidate sentences against each token in the reference sentences, using contextual embeddings derived from pre-trained BERT-based models. From these similarities, BERTScore computes precision, recall, and F1 scores. In our metrics, we utilize the precision and F1 scores, employing DeBERTa-XL-MNLI~\cite{he2020deberta} as our model, as recommended by the authors.

\paragraph{YISI-1 \cite{lo2019yisi}} The metric computes the semantic similarity between translations from MT and human references by aggregating lexical semantic similarities, which are weighted by inverse document frequency (IDF) based on the contextual embeddings extracted from pre-trained language model, specifically the last hidden layer of mBERT in our case.

\paragraph{BLEURT \cite{sellam2020bleurt}} The metric is fine-tuned using Direct Assessment (DA) dataset. BLEURT jointly encodes the translation and reference using the \texttt{[CLS]} token as an embedding to represent the pair. We employ the BLEURT-20 checkpoint \cite{pu2021learning}, which was trained on RemBERT \cite{chung2020rethinking} using DA data from prior shared tasks between 2015 and 2019 and augmented with synthetic data generated from Wikipedia articles.

\paragraph{COMET-22 \cite{rei2022comet}} The metric is an ensemble of COMET estimator \cite{rei2020comet} fine-tuned on DA and a Sequence Tagger trained on Multidimensional Quality Metrics (MQM) annotations. We utilize the \texttt{wmt22-comet-da} as our COMET-22 checkpoint, in which the COMET Estimator model and the sequence tagging model are trained on top of XLM-R using DA from 2017 to 2020 and InfoXLM \cite{chi2021infoxlm}, respectively.

\paragraph{XCOMET-XL \cite{guerreiro2023xcomet}} The metric that performs both sentence-level evaluation and error span detection, making it a more interpretable learned metric. The model utilizes XLM-R XL (3.5B)~\cite{goyal2021larger} which is trained in stages, starting with DA annotations and then fine-tuned on MQM data.

\paragraph{MetricX-23 \cite{juraska2023metricx}} The metric uses mT5 encoder-decoder language model. We leverage three different variations of MetricX-23, each fine-tuned from the mT5-Large, mT5-XL, and mT5-XXL respectively. The fine-tuning was performed using DA data from 2015-2020, MQM data from 2020-2021, and synthetic data.

\subsubsection{Reference-free Metric}
We utilize six different metrics in our optimization, including two variations of CometKiwi and three variations of MetricX-23. We describe the reference-free metrics used for \benchmarkname-QE as follows:

\paragraph{CometKiwi \cite{rei2022comet}} The metric is a reference-free learned metric fine-tuned on DA on top of RemBERT \cite{chung2020rethinking} and the same sequence tagger as COMET-22. However, it operates with reference-less inputs during inference. We use two distinct metrics from CometKiwi, each associated with its own separate checkpoint: \texttt{wmt22-cometkiwi-da} and \texttt{wmt23-cometkiwi-da-xl}. The latter checkpoint replaces InfoXLM with XLM-R XL (3.5B) and is trained on the same dataset, but it also includes newly released DA for Indian languages, which were added as additional training data for the 2023 Quality Estimation (QE) shared task.

\paragraph{GEMBA-MQM \cite{kocmi2023gemba}} The metric is a GPT-based evaluation metric designed for error quality span marking. It employs a three-shot prompting approach using the GPT-4 model, specifically GPT-4o mini in our case.

\paragraph{MetricX-23-QE \cite{juraska2023metricx}} The metric is a reference-free learned metric similar to MetricX-23. We also utilize three different variations, each fine-tuned from the mT5-L, mT5-XL, and mT5-XXL checkpoints, respectively.

\begin{table}[!th]
\centering
\resizebox{0.49\textwidth}{!}{
\begin{tabular}{lccc}
\toprule
\textbf{Metric} & \textbf{all} & \textbf{sys} & \textbf{seg} \\
\midrule
\textbf{Reference-based} \\ \midrule
sentinel-ref-mqm & 0.513 & 0.510 & 0.515 \\
BLEU & 0.589 & 0.663 & 0.515 \\
spBLEU & 0.593 & 0.669 & 0.516 \\
chrF & 0.606 & 0.693 & 0.520 \\
chrfS & 0.608 & 0.699 & 0.516 \\
BERTScore & 0.609 & 0.697 & 0.522 \\
MEE4 & 0.617 & 0.712 & 0.522 \\
damonmonli & 0.640 & 0.734 & 0.547 \\
YiSi-1 & 0.642 & 0.760 & 0.524 \\
PrismRefSmall & 0.646 & 0.766 & 0.526 \\
PrismRefMedium & 0.650 & 0.739 & 0.560 \\
BLCOM\_1 & 0.684 & 0.803 & 0.566 \\
BLEURT-20 & 0.686 & 0.821 & 0.550 \\
COMET-22 & 0.695 & 0.833 & 0.557 \\
XCOMET & 0.719 & \textbf{0.862} & 0.576 \\
MetricX-24 (Hybrid) & \underline{0.721} & \underline{0.857} & \underline{0.586} \\ \midrule
$\benchmarkname$ (Hybrid) & \textbf{0.725} & 0.853 & \textbf{0.596} \\ \midrule
\textbf{Reference-free} \\ \midrule
sentinel-src-mqm & 0.513 & 0.511 & 0.515 \\
XLsimMqm & 0.515 & 0.506 & 0.523 \\
sentinel-cand-mqm & 0.630 & 0.734 & 0.525 \\
CometKiwi & 0.635 & 0.738 & 0.532 \\
bright-qe & 0.664 & 0.788 & 0.541 \\
XCOMET-QE & 0.688 & 0.823 & 0.554 \\
gemba\_esa & \underline{0.711} & \underline{0.846} & \underline{0.576} \\
MetricX-24-QE (Hybrid) & \textbf{0.714} & \textbf{0.847} & \textbf{0.580} \\ \midrule
$\benchmarkname$-\textsc{QE} & 0.681 & 0.804 & 0.557 \\
\bottomrule
\end{tabular}
}
\caption{Detailed WMT24 results for segment-level and system-level. \textbf{Bold} and \underline{underline} values indicate the best and second best performance, respectively.}
\label{tab:seg-sys-result}
\end{table}

\begin{figure}[!th]
    \centering
    \begin{subfigure}[!t]{0.49\textwidth}
        \centering
        \caption{}
        \includegraphics[width=\linewidth]{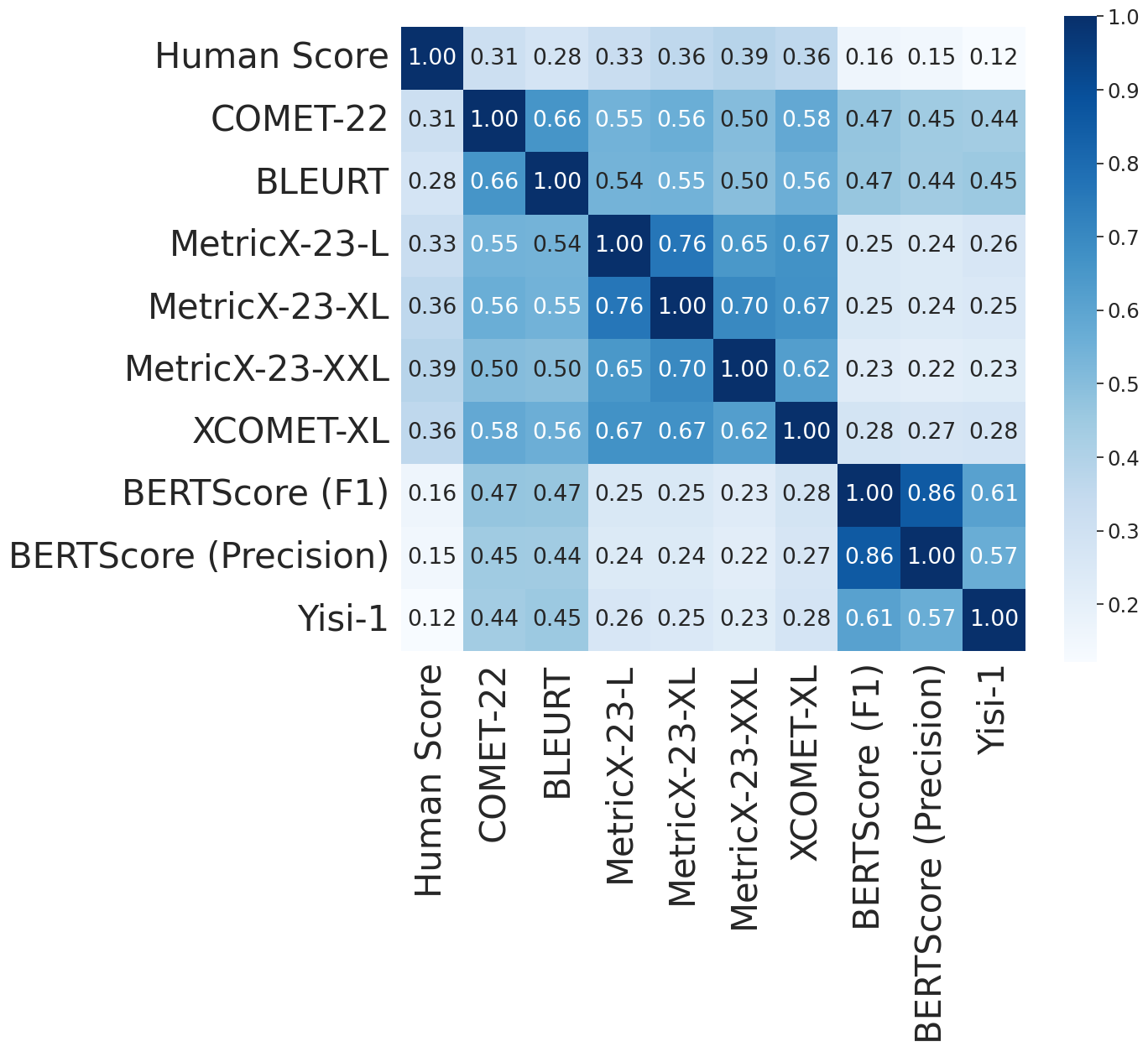} 
        
    \end{subfigure}
    \begin{subfigure}[!t]{0.49\textwidth}
        \centering
        \caption{}
        \includegraphics[width=\linewidth]{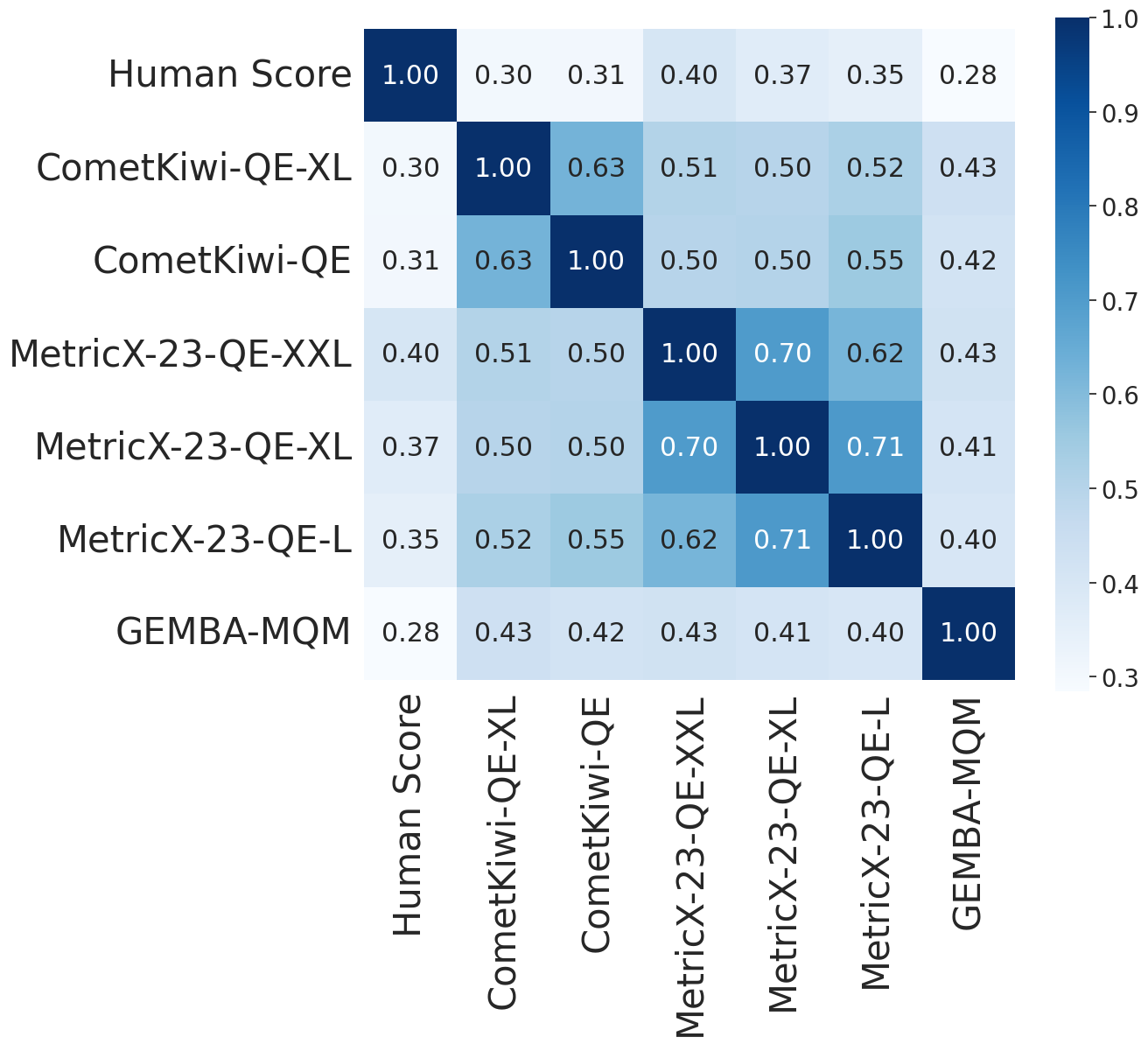} 
        
    \end{subfigure}
    \caption{Heatmaps showing Kendall correlation coefficients between human scores and MT metrics over 3 years of MQM datasets from the WMT shared tasks (2020-2022). Panel (a) displays correlations for the metrics used in $\benchmarkname$, while panel (b) displays correlations for the metrics used in $\benchmarkname$-QE.}
    \label{fig:heatmaps}
\end{figure}

\section{Results and Discussion}

\subsection{Optimized Metric Configuration}
Table~\ref{tab:best-configs} shows the weight proportion of each metric for $\benchmarkname$. The optimized configuration is notably sparse. When a metric does not positively contribute to improving performance, the GP assigns it a weight of zero. This is supported by Figure~\ref{fig:heatmaps}, where the GP selects metrics with high Kendall correlation coefficients relative to other provided metrics. In contrast, metrics with low Kendall correlation coefficients are excluded.

Interestingly, in both reference-based and reference-free settings, the optimization process consistently selects only one variant of MetricX-23, specifically MetricX-23-XXL, even though all three variants of MetricX-23 exhibit high Kendall correlation coefficients. The optimization process favors MetricX-23-XXL as the highest-performing metric, leading to the exclusion of the other two variants during the GP assignment. This enhances the efficiency of $\benchmarkname$ as we would only need to use fewer metrics for $\benchmarkname$. Thus, given a set of metrics, the optimization process would prioritize high-performing metrics, such as the MetricX-23 and COMET variants as shown, leading $\benchmarkname$ and $\benchmarkname$-QE to construct a better and more robust metric.

\subsection{Results on WMT24 Shared Task}
Table~\ref{tab:qa-results} presents the WMT24 shared task results, including system-level soft pairwise ranking accuracy (sys SPA) proposed by \citet{thompson2024improving}, segment-level pairwise ranking accuracy with tie calibration (seg acc-t) as described by \citet{deutsch2023ties}, and system- and segment-level Pearson correlation (avg. corr), as outlined in the WMT23 Metrics Shared Task \cite{freitag2023results}. Based on the overall system and segment average correlation and system accuracy, $\benchmarkname$ outperforms all metrics in the primary submission, with $\benchmarkname$ (Hybrid) achieving the highest performance among its variants.

Table \ref{tab:lang-cat-result} further highlights the performance, where $\benchmarkname$ delivers superior results for en-es, while also maintaining strong performance in en-de and ja-zh, indicating that our methods generalize well across different language pairs. The breakdown in Table \ref{tab:seg-sys-result} shows that $\benchmarkname$ achieves the best segment-level performance, consistent with our optimization approach targeting Kendall correlation at the segment level. Given that our metric optimization focuses solely on segment-level correlation, incorporating a different weighting method to account for system-level settings could further improve $\benchmarkname$'s alignment with system-level accuracy. While $\benchmarkname$-\textsc{QE} does not match the performance of gemba\_esa, MetricX-24-QE (Hybrid), or CometKiwi-XXL, it remains competitive at the segment level for the en-es language pair. Incorporating better reference-free models such as CometKiwi-XXL and GEMBA-MQM with GPT-4o instead of GPT-4o mini may help improve the performance of $\benchmarkname$-\textsc{QE}.

\subsection{Compute Efficiency}
We only run models that can be executed on GPUs with 40GB of memory. We limit our resource usage to GPT-4o mini, a smaller and lower-performing version of GPT-4o, while GEMBA-MQM is a GPT-4 based metric. This constraint restricts our ability to achieve state-of-the-art results or surpass GEMBA-based metrics using GPT-4. However, we demonstrate that even without employing high-memory models like XCOMET-Ensemble in our reference-based setting, we can still outperform other models. Additionally, our QE metric remains competitive and on par with XCOMET-QE.

\section{Conclusion}
In this paper, we propose $\benchmarkname$, a novel metric designed to evaluate MT tasks by aligning with human preferences through Bayesian optimization with GP. $\benchmarkname$ effectively combines and optimizes existing MT metrics based on human feedback, resulting in a highly flexible and efficient evaluation tool. Our findings show that $\benchmarkname$ surpasses existing baselines for reference-based metrics, establishing a new state-of-the-art, while its reference-free metric performance rivals the best models available. Additionally, $\benchmarkname$ can be tailored to various factors, such as performance and efficiency, making it adaptable to diverse requirements.

\section*{Ethical Considerations}
Our research focuses on evaluating MT systems using a newly proposed metric. We are committed to conducting our evaluations with the highest levels of transparency and fairness. By prioritizing these principles, we aim to set a standard for reliability and objectivity in the assessment of the system.

\section*{Limitations}
We optimize $\benchmarkname$ using segment-level scores from the MQM dataset. Future work could extend this to other objective functions or system-level optimization and explore non-MQM datasets like DA for further insights. We did not include metrics such as XCOMET-XXL, XCOMET-Ensemble, and XCOMET-QE-Ensemble due to computational constraints.

\bibliography{custom}
\bibliographystyle{acl_natbib}

\appendix

\section{Pre-processing}
\label{sec-preprocessing}
The pre-processing can be defined as follows:
\begin{enumerate}
    \item \textbf{Clipping:} Let the valid range for \(y_i\) be defined by \([y_{i}^{\text{min}}, y_{i}^{\text{max}}]\). The clipped metric score $y_i'$ can be defined as:
    \begin{align}
    y_i' = 
    \begin{cases} 
    y_{i}^{\text{min}} & \text{if } y_i < y_{i}^{\text{min}}, \\
    y_i & \text{if } y_{i}^{\text{min}} \leq y_i \leq y_{i}^{\text{max}}, \\
    y_{i}^{\text{max}} & \text{if } y_i > y_{i}^{\text{max}}.
    \end{cases}
    \end{align}
    \item \textbf{Normalization:} After clipping, the score is normalized to a common scale of \([0, 1]\):
    \begin{align}
    \tilde{y}_i = \frac{y_i' - y_{i}^{\text{min}}}{y_{i}^{\text{max}} - y_{i}^{\text{min}}}.
    \end{align}
    \item \textbf{Inversion (if applicable):} If the metric is such that higher scores indicate worse performance, we invert the normalized score:
    \begin{align}
    \tilde{y}_i = 1 - \tilde{y}_i.
    \end{align}
\end{enumerate}

\section{Additional Results}
\label{sec:appendix}

We provide additional details for the results of WMT24 for each task in Tables~\ref{tab:en-de-results},~\ref{tab:en-es-results}, and~\ref{tab:ja-zh-results}. Additional results for each domain are also provided in Table~\ref{tab:result-domain}.

\begin{table*}[!th]
\centering
    \resizebox{0.95\textwidth}{!}{
    \begin{tabular}{lcc|cc|cc|cc|cc|cc|cc|cc}
        \toprule
        \textbf{Domain} & \multicolumn{2}{c|}{\textbf{literary}} & \multicolumn{2}{c|}{\textbf{news}} & \multicolumn{2}{c|}{\textbf{social}} & \multicolumn{2}{c|}{\textbf{speech}} & \multicolumn{2}{c|}{\textbf{literary}} & \multicolumn{2}{c|}{\textbf{news}} & \multicolumn{2}{c|}{\textbf{social}} & \multicolumn{2}{c@{}}{\textbf{speech}} \\
        \textbf{Metric} & \multicolumn{2}{c|}{task1} & \multicolumn{2}{c|}{task2} & \multicolumn{2}{c|}{task3} & \multicolumn{2}{c|}{task4} & \multicolumn{2}{c|}{task5} & \multicolumn{2}{c|}{task6} & \multicolumn{2}{c|}{task7} & \multicolumn{2}{c@{}}{task8} \\
        \textbf{Level} & r & sys SPA & r & sys SPA & r & sys SPA & r & sys SPA & r & seg acc-t & r & seg acc-t & r & seg acc-t & r & seg acc-t \\
        \midrule
        \textbf{Reference-based} \\ \midrule
        sentinel-ref-mqm & 4 & 0.525 & 4 & 0.535 & 6 & 0.439 & 6 & 0.461 & 9 & 0.351 & 9 & 0.421 & 16 & 0.520 & 13 & 0.240 \\
        BLEU & 2 & 0.795 & 1 & 0.807 & 5 & 0.691 & 4 & 0.709 & 5 & 0.535 & 9 & 0.421 & 15 & 0.522 & 11 & 0.433 \\
        spBLEU & 2 & 0.785 & 1 & 0.810 & 5 & 0.697 & 4 & 0.700 & 4 & 0.540 & 9 & 0.421 & 15 & 0.522 & 10 & 0.446 \\
        chrF & 2 & 0.774 & 1 & \textbf{0.831} & 4 & 0.728 & 3 & 0.723 & 4 & 0.540 & 9 & 0.421 & 14 & 0.523 & 10 & 0.445 \\
        chrfS & 2 & 0.797 & 1 & 0.826 & 4 & 0.712 & 3 & 0.736 & 4 & 0.543 & 9 & 0.421 & 13 & 0.525 & 9 & 0.449 \\
        BERTScore & 2 & 0.777 & 1 & 0.821 & 4 & 0.708 & 4 & 0.712 & 4 & 0.550 & 8 & 0.424 & 12 & 0.526 & 11 & 0.436 \\
        MEE4 & 2 & 0.792 & 1 & 0.826 & 5 & 0.688 & 4 & 0.712 & 4 & 0.549 & 9 & 0.421 & 10 & 0.531 & 9 & 0.452 \\
        damonmonli & 2 & 0.734 & 1 & 0.788 & 5 & 0.695 & 5 & 0.613 & 7 & 0.503 & 7 & 0.427 & 14 & 0.523 & 12 & 0.404 \\
        YiSi-1 & 2 & 0.761 & 1 & 0.822 & 4 & 0.719 & 3 & 0.760 & 3 & 0.555 & 9 & 0.421 & 12 & 0.526 & 8 & 0.456 \\
        PrismRefSmall & 2 & 0.786 & 1 & \underline{0.829} & 4 & 0.750 & 3 & 0.736 & 5 & 0.526 & 8 & 0.423 & 13 & 0.524 & 7 & 0.464 \\
        PrismRefMedium & 2 & 0.761 & 1 & \textbf{0.831} & 3 & 0.756 & 4 & 0.722 & 4 & 0.536 & 8 & 0.424 & 11 & 0.528 & 8 & 0.461 \\
        BLCOM\_1 & 1 & 0.828 & 1 & 0.812 & 3 & 0.803 & 2 & 0.833 & 3 & 0.562 & 7 & 0.427 & 9 & 0.535 & 5 & 0.487 \\
        BLEURT-20 & 1 & 0.827 & 2 & 0.768 & 2 & 0.842 & 3 & 0.784 & 4 & 0.544 & 5 & 0.444 & 7 & 0.554 & 4 & 0.494 \\
        COMET-22 & 1 & 0.814 & 1 & 0.804 & 2 & 0.852 & 2 & 0.813 & 2 & 0.571 & 6 & 0.437 & 6 & 0.559 & 3 & 0.503 \\
        XCOMET & 1 & \underline{0.830} & 1 & 0.782 & 1 & \underline{0.889} & 2 & \textbf{0.845} & 2 & 0.573 & 3 & \underline{0.479} & 2 & 0.575 & 2 & \underline{0.510} \\
        MetricX-24 (Hybrid) & 1 & \textbf{0.840} & 1 & 0.774 & 1 & 0.874 & 2 & \underline{0.816} & 2 & \underline{0.580} & 3 & 0.478 & 2 & \underline{0.576} & 1 & \textbf{0.520} \\ \midrule
        $\benchmarkname$ (Hybrid) & 1 & 0.822 & 2 & 0.763 & 1 & \textbf{0.896} & 3 & 0.788 & 1 & \textbf{0.597} & 2 & \textbf{0.493} & 1 & \textbf{0.588} & 2 & 0.506 \\ \midrule
        \textbf{Reference-free} \\ \midrule
        sentinel-src-mqm & 4 & 0.525 & 4 & 0.534 & 6 & 0.438 & 6 & 0.461 & 9 & 0.351 & 9 & 0.421 & 16 & 0.520 & 13 & 0.240 \\
        XLsimMqm & 4 & 0.478 & 4 & 0.497 & 5 & 0.613 & 3 & 0.768 & 8 & 0.474 & 1 & \textbf{0.532} & 10 & 0.531 & 12 & 0.410 \\
        sentinel-cand-mqm & 2 & 0.776 & 2 & 0.735 & 1 & \textbf{0.896} & 3 & 0.760 & 4 & 0.547 & 2 & \underline{0.501} & 4 & 0.569 & 6 & 0.480 \\
        CometKiwi & 3 & 0.722 & 2 & 0.723 & 4 & 0.732 & 4 & 0.685 & 5 & 0.535 & 5 & 0.445 & 9 & 0.532 & 10 & 0.443 \\
        bright-qe & 2 & 0.795 & 2 & 0.755 & 3 & 0.760 & 2 & 0.827 & 6 & 0.517 & 4 & 0.457 & 8 & 0.547 & 7 & 0.469 \\
        XCOMET-\textsc{QE} & 2 & 0.758 & 1 & \textbf{0.790} & 2 & 0.850 & 1 & \textbf{0.882} & 4 & 0.541 & 3 & 0.480 & 5 & 0.565 & 3 & \underline{0.498} \\
        gemba\_esa & 1 & \textbf{0.820} & 2 & 0.755 & 3 & 0.801 & 2 & 0.815 & 3 & \underline{0.562} & 5 & 0.450 & 3 & \underline{0.569} & 6 & 0.474 \\
        MetricX-24-QE (Hybrid) & 2 & \underline{0.809} & 1 & \underline{0.783} & 1 & \underline{0.863} & 1 & \underline{0.860} & 2 & \textbf{0.575} & 4 & 0.460 & 3 & \textbf{0.573} & 1 & \textbf{0.518} \\ \midrule
        $\benchmarkname$-\textsc{QE} & 3 & 0.691 & 3 & 0.690 & 2 & 0.811 & 1 & 0.852 & 6 & 0.520 & 4 & 0.457 & 6 & 0.555 & 7 & 0.471 \\
        \bottomrule
    \end{tabular}
    }
\caption{Detailed result for language pair en-de. \textbf{Bold} and \underline{underline} values indicate the best and second best performance, respectively.}
\label{tab:en-de-results}
\end{table*}

\begin{table*}[!th]
\centering
    \resizebox{0.95\textwidth}{!}{
    \begin{tabular}{lcc|cc|cc|cc|cc|cc|cc|cc}
        \toprule
        \textbf{Domain} & \multicolumn{2}{c|}{\textbf{literary}} & \multicolumn{2}{c|}{\textbf{news}} & \multicolumn{2}{c|}{\textbf{social}} & \multicolumn{2}{c|}{\textbf{speech}} & \multicolumn{2}{c|}{\textbf{literary}} & \multicolumn{2}{c|}{\textbf{news}} & \multicolumn{2}{c|}{\textbf{social}} & \multicolumn{2}{c@{}}{\textbf{speech}} \\
        \textbf{Metric} & \multicolumn{2}{c|}{task9} & \multicolumn{2}{c|}{task10} & \multicolumn{2}{c|}{task11} & \multicolumn{2}{c|}{task12} & \multicolumn{2}{c|}{task13} & \multicolumn{2}{c|}{task14} & \multicolumn{2}{c|}{task15} & \multicolumn{2}{c@{}}{task16} \\
        \textbf{Level} & r & sys SPA & r & sys SPA & r & sys SPA & r & sys SPA & r & seg acc-t & r & seg acc-t & r & seg acc-t & r & seg acc-t \\
        \midrule
        \textbf{Reference-based} \\ \midrule
        sentinel-ref-mqm & 3 & 0.564 & 4 & 0.460 & 5 & 0.599 & 4 & 0.556 & 5 & 0.615 & 4 & 0.715 & 8 & 0.744 & 6 & 0.535 \\
        BLEU & 3 & 0.595 & 4 & 0.557 & 5 & 0.624 & 5 & 0.480 & 5 & 0.615 & 4 & 0.715 & 7 & 0.745 & 5 & 0.536 \\
        spBLEU & 3 & 0.602 & 3 & 0.595 & 4 & 0.635 & 5 & 0.486 & 4 & 0.615 & 4 & 0.715 & 7 & 0.745 & 5 & 0.536 \\
        chrF & 3 & 0.621 & 3 & 0.593 & 4 & 0.657 & 5 & 0.490 & 4 & 0.615 & 4 & 0.715 & 8 & 0.744 & 4 & 0.537 \\
        chrfS & 2 & 0.648 & 3 & 0.604 & 4 & 0.667 & 5 & 0.472 & 3 & 0.617 & 4 & 0.715 & 6 & 0.746 & 5 & 0.537 \\
        BERTScore & 2 & 0.665 & 1 & 0.715 & 3 & 0.679 & 5 & 0.488 & 3 & 0.617 & 2 & \underline{0.717} & 5 & 0.747 & 5 & 0.537 \\
        MEE4 & 2 & 0.651 & 2 & 0.628 & 3 & 0.677 & 5 & 0.467 & 3 & 0.617 & 4 & 0.715 & 3 & 0.750 & 4 & 0.539 \\
        damonmonli & 1 & 0.720 & 2 & 0.673 & 2 & 0.737 & 4 & 0.555 & 2 & 0.621 & 4 & 0.715 & 5 & 0.747 & 5 & 0.536 \\
        YiSi-1 & 1 & 0.706 & 2 & 0.673 & 3 & \underline{0.715} & 5 & 0.505 & 3 & 0.617 & 4 & 0.715 & 6 & 0.745 & 4 & 0.538 \\
        PrismRefSmall & 1 & 0.727 & 2 & 0.624 & 2 & 0.724 & 5 & 0.518 & 5 & 0.615 & 3 & 0.716 & 8 & 0.745 & 5 & 0.537 \\
        PrismRefMedium & 1 & 0.733 & 2 & 0.649 & 2 & 0.745 & 5 & 0.518 & 4 & 0.616 & 3 & 0.716 & 7 & 0.745 & 5 & 0.536 \\
        BLCOM\_1 & 2 & 0.702 & 2 & 0.675 & 2 & 0.773 & 4 & 0.623 & 3 & 0.617 & 4 & 0.715 & 5 & 0.747 & 4 & 0.541 \\
        BLEURT-20 & 2 & 0.702 & 2 & 0.648 & 1 & 0.841 & 4 & 0.587 & 2 & 0.620 & 4 & 0.715 & 6 & 0.746 & 6 & 0.535 \\
        COMET-22 & 1 & \textbf{0.755} & 1 & \textbf{0.731} & 1 & \textbf{0.865} & 3 & 0.653 & 1 & \textbf{0.626} & 4 & 0.715 & 4 & 0.750 & 3 & \underline{0.551} \\
        XCOMET & 1 & 0.733 & 1 & 0.677 & 1 & 0.840 & 2 & \underline{0.685} & 1 & \underline{0.625} & 2 & \underline{0.717} & 1 & \textbf{0.756} & 3 & 0.548 \\
        MetricX-24 (Hybrid) & 1 & \underline{0.741} & 1 & 0.683 & 1 & 0.846 & 2 & \textbf{0.691} & 2 & 0.621 & 4 & 0.715 & 3 & 0.750 & 2 & \textbf{0.559} \\ \midrule
        $\benchmarkname$ (Hybrid) & 1 & 0.734 & 1 & 0.688 & 1 & \underline{0.852} & 2 & 0.682 & 2 & 0.619 & 1 & \textbf{0.720} & 2 & \underline{0.753} & 3 & 0.550 \\ \midrule
        \textbf{Reference-free} \\ \midrule
        sentinel-src-mqm & 3 & 0.565 & 4 & 0.456 & 5 & 0.598 & 4 & 0.554 & 5 & 0.615 & 4 & 0.715 & 8 & 0.744 & 6 & 0.535 \\
        XLsimMqm & 4 & 0.363 & 2 & 0.645 & 6 & 0.410 & 3 & 0.640 & 4 & 0.615 & 4 & 0.715 & 6 & 0.745 & 4 & 0.537 \\
        sentinel-cand-mqm & 2 & 0.695 & 1 & 0.678 & 2 & 0.780 & 2 & 0.690 & 2 & 0.620 & 1 & \underline{0.720} & 4 & 0.749 & 4 & 0.537 \\
        CometKiwi & 2 & 0.641 & 2 & 0.661 & 2 & 0.767 & 2 & 0.681 & 2 & 0.620 & 3 & 0.716 & 3 & \underline{0.751} & 3 & 0.553 \\
        bright-qe & 3 & 0.583 & 1 & 0.677 & 2 & 0.764 & 1 & \textbf{0.772} & 2 & 0.621 & 2 & 0.718 & 3 & \underline{0.751} & 1 & \textbf{0.571} \\
        XCOMET-\textsc{QE} & 1 & \underline{0.731} & 2 & 0.673 & 2 & 0.779 & 2 & 0.700 & 2 & \underline{0.622} & 1 & \textbf{0.721} & 2 & \textbf{0.754} & 3 & 0.547 \\
        gemba\_esa & 1 & \textbf{0.740} & 1 & \textbf{0.723} & 1 & \textbf{0.820} & 2 & \underline{0.704} & 2 & 0.621 & 2 & 0.718 & 5 & 0.746 & 3 & 0.549 \\
        MetricX-24-\textsc{QE} (Hybrid) & 1 & 0.727 & 1 & 0.694 & 1 & \underline{0.818} & 2 & 0.703 & 1 & \underline{0.622} & 4 & 0.715 & 5 & 0.748 & 2 & 0.563 \\ \midrule
        $\benchmarkname$-\textsc{QE} & 2 & 0.661 & 1 & \underline{0.711} & 2 & 0.751 & 2 & 0.692 & 1 & \textbf{0.624} & 2 & 0.717 & 4 & 0.749 & 2 & \underline{0.565} \\
        \bottomrule
    \end{tabular}
    }
\caption{Detailed WMT24 result for language pair en-es. \textbf{Bold} and \underline{underline} values indicate the best and second best performance, respectively.}
\label{tab:en-es-results}
\end{table*}

\begin{table*}[!th]
\centering
    \resizebox{0.95\textwidth}{!}{
    \begin{tabular}{lcc|cc|cc|cc|cc|cc}
        \toprule
        \textbf{Domain} & \multicolumn{2}{c|}{\textbf{literary}} & \multicolumn{2}{c|}{\textbf{news}} & \multicolumn{2}{c|}{\textbf{speech}} & \multicolumn{2}{c|}{\textbf{literary}} & \multicolumn{2}{c|}{\textbf{news}} & \multicolumn{2}{c@{}}{\textbf{speech}} \\
        \textbf{Metric} & \multicolumn{2}{c|}{task17} & \multicolumn{2}{c|}{task18} & \multicolumn{2}{c|}{task19} & \multicolumn{2}{c|}{task20} & \multicolumn{2}{c|}{task21} & \multicolumn{2}{c@{}}{task22} \\
        \textbf{Level} & r & sys SPA & r & sys SPA & r & sys SPA & r & seg acc-t & r & seg acc-t & r & seg acc-t \\
        \midrule
\textbf{Reference-based} \\ \midrule
        sentinel-ref-mqm & 5 & 0.504 & 7 & 0.494 & 8 & 0.569 & 11 & 0.532 & 8 & 0.497 & 12 & 0.197 \\
        BLEU & 4 & 0.637 & 3 & 0.762 & 8 & 0.562 & 11 & 0.532 & 8 & 0.497 & 11 & 0.205 \\
        spBLEU & 4 & 0.699 & 4 & 0.755 & 7 & 0.743 & 9 & 0.535 & 7 & 0.497 & 7 & 0.506 \\
        chrF & 4 & 0.721 & 3 & 0.768 & 6 & 0.766 & 9 & 0.536 & 8 & 0.497 & 6 & 0.513 \\
        chrfS & 3 & 0.768 & 3 & 0.773 & 5 & 0.823 & 9 & 0.537 & 7 & 0.497 & 5 & 0.526 \\
        BERTScore & 3 & 0.786 & 5 & 0.748 & 5 & 0.833 & 9 & 0.536 & 6 & 0.500 & 5 & 0.524 \\
        MEE4 & 2 & 0.816 & 3 & 0.789 & 2 & 0.892 & 8 & 0.538 & 7 & 0.497 & 5 & 0.521 \\
        damonmonli & 2 & 0.839 & 1 & \textbf{0.857} & 2 & 0.893 & 7 & 0.545 & 5 & 0.504 & 8 & 0.495 \\
        YiSi-1 & 2 & 0.813 & 4 & 0.758 & 4 & 0.853 & 8 & 0.539 & 6 & 0.502 & 4 & 0.535 \\
        PrismRefSmall & 2 & 0.850 & 3 & 0.786 & 4 & 0.854 & 11 & 0.532 & 7 & 0.498 & 3 & 0.541 \\
        PrismRefMedium & 2 & 0.839 & 3 & 0.794 & 3 & 0.875 & 11 & 0.532 & 7 & 0.499 & 3 & 0.544 \\
        BLCOM\_1 & 2 & 0.827 & 3 & 0.779 & 1 & \underline{0.909} & 7 & 0.545 & 6 & 0.500 & 1 & \underline{0.552} \\
        BLEURT-20 & 1 & 0.864 & 3 & 0.797 & 2 & 0.904 & 8 & 0.539 & 5 & 0.508 & 4 & 0.535 \\
        COMET-22 & 2 & 0.811 & 5 & 0.714 & 2 & 0.906 & 6 & 0.557 & 4 & 0.517 & 1 & \underline{0.552} \\
        XCOMET & 2 & 0.850 & 2 & \underline{0.832} & 1 & \textbf{0.924} & 5 & 0.566 & 3 & 0.527 & 1 & \textbf{0.558} \\
        MetricX-24 (Hybrid) & 1 & \textbf{0.893} & 3 & 0.804 & 3 & 0.890 & 2 & \underline{0.607} & 2 & \underline{0.543} & 2 & 0.551 \\ \midrule
        $\benchmarkname$ (Hybrid) & 1 & \underline{0.876} & 3 & 0.805 & 2 & 0.896 & 1 & \textbf{0.643} & 1 & \textbf{0.551} & 3 & 0.544 \\ \midrule
        \textbf{Reference-free} \\ \midrule
        sentinel-src-mqm & 5 & 0.522 & 7 & 0.491 & 8 & 0.570 & 11 & 0.532 & 8 & 0.497 & 12 & 0.197 \\
        XLsimMqm & 5 & 0.592 & 6 & 0.506 & 8 & 0.574 & 9 & 0.535 & 8 & 0.497 & 10 & 0.420 \\
        sentinel-cand-mqm & 5 & 0.595 & 6 & 0.581 & 7 & 0.681 & 5 & 0.565 & 5 & 0.505 & 9 & 0.445 \\
        CometKiwi & 4 & 0.667 & 3 & 0.797 & 4 & 0.858 & 7 & 0.549 & 4 & 0.519 & 6 & 0.516 \\
        bright-qe & 3 & 0.738 & 3 & 0.786 & 6 & 0.759 & 7 & 0.547 & 3 & 0.528 & 9 & 0.438 \\
        XCOMET-QE & 3 & 0.740 & 3 & 0.806 & 3 & 0.868 & 10 & 0.534 & 5 & 0.504 & 6 & 0.514 \\
        gemba\_esa & 2 & \underline{0.832} & 1 & \textbf{0.882} & 1 & \textbf{0.930} & 3 & \underline{0.592} & 2 & \textbf{0.545} & 3 & \underline{0.538} \\
        MetricX-24-\textsc{QE} (Hybrid) & 2 & \textbf{0.853} & 2 & \underline{0.814} & 2 & \underline{0.907} & 3 & \textbf{0.597} & 3 & \underline{0.529} & 2 & \textbf{0.548} \\ \midrule
        $\benchmarkname$-\textsc{QE} & 3 & 0.768 & 4 & 0.749 & 3 & 0.878 & 4 & 0.585 & 4 & 0.522 & 7 & 0.505 \\
        \bottomrule
    \end{tabular}
    }
\caption{Detailed WMT24 result for language pair ja-zh. \textbf{Bold} and \underline{underline} values indicate the best and second best performance, respectively.}
\label{tab:ja-zh-results}
\end{table*}

\begin{table*}[!ht]
\centering
\resizebox{0.95\textwidth}{!}{
\begin{tabular}{lccccc}
\toprule
\textbf{Metric} & \textbf{all} & \textbf{literary} & \textbf{news} & \textbf{social} & \textbf{speech} \\
& & Task 1,5,9,13,17,20 & Task 2,6,10,14,18,21 & Task 3,7,11,16 & Task 4,8,12,16,19,22 \\
\midrule
\textbf{Reference-based} \\ \midrule
sentinel-ref-mqm & 0.513 & 0.515 & 0.520 & 0.576 & 0.426 \\
BLEU & 0.589 & 0.618 & 0.626 & 0.645 & 0.488 \\
spBLEU & 0.593 & 0.629 & 0.632 & 0.650 & 0.570 \\
chrF & 0.606 & 0.635 & 0.637 & 0.663 & 0.579 \\
chrfS & 0.608 & 0.652 & 0.640 & 0.662 & 0.590 \\
BERTScore & 0.609 & 0.655 & 0.654 & 0.665 & 0.588 \\
MEE4 & 0.617 & 0.661 & 0.646 & 0.661 & 0.598 \\
damonmonli & 0.640 & 0.660 & 0.661 & 0.676 & 0.583 \\
YiSi-1 & 0.642 & 0.665 & 0.649 & 0.676 & 0.608 \\
PrismRefSmall & 0.646 & 0.672 & 0.646 & 0.686 & 0.608 \\
PrismRefMedium & 0.650 & 0.669 & 0.652 & 0.693 & 0.609 \\
BLCOM\_1 & 0.684 & 0.680 & 0.651 & 0.714 & 0.658 \\
BLEURT-20 & 0.686 & 0.683 & 0.647 & 0.746 & 0.640 \\
COMET-22 & 0.695 & 0.689 & 0.653 & 0.757 & 0.663 \\
XCOMET & 0.719 & 0.696 & \underline{0.669} & 0.765 & \textbf{0.678} \\
MetricX-24 (Hybrid) & \underline{0.721} & \underline{0.714} & 0.666 & \underline{0.761} & \underline{0.671} \\ \midrule
$\benchmarkname$ (Hybrid) & \textbf{0.725} & \textbf{0.715} & \textbf{0.670} & \textbf{0.772} & 0.661 \\ \midrule
\textbf{Reference-free} \\ \midrule
sentinel-src-mqm & 0.513 & 0.518 & 0.519 & 0.575 & 0.426 \\
XLsimMqm & 0.515 & 0.509 & 0.565 & 0.575 & 0.558 \\
sentinel-cand-mqm & 0.630 & 0.633 & 0.620 & \underline{0.748} & 0.599 \\
CometKiwi & 0.635 & 0.622 & 0.643 & 0.695 & 0.623 \\
bright-qe & 0.664 & 0.634 & 0.653 & 0.706 & 0.639 \\
XCOMET-QE & 0.688 & 0.654 & 0.662 & 0.737 & \underline{0.668} \\
gemba\_esa & \underline{0.711} & \underline{0.694} & \textbf{0.679} & 0.734 & \underline{0.668} \\
MetricX-24-Hybrid-QE & \textbf{0.714} & \textbf{0.697} & \underline{0.666} & \textbf{0.751} & \textbf{0.683} \\ \midrule
$\benchmarkname$-\textsc{QE} & 0.681 & 0.641 & 0.641 & 0.717 & 0.660 \\
\bottomrule
\end{tabular}
}
\caption{Detailed WMT24 results per domain category. \textbf{Bold} and \underline{underline} values indicate the best and second best performance, respectively.}
\label{tab:result-domain}
\end{table*}

\end{document}